\documentclass[11pt]{article}

\usepackage[final]{acl}

\usepackage{times}
\usepackage{latexsym}

\usepackage[T1]{fontenc}

\usepackage[utf8]{inputenc}

\usepackage{microtype}

\usepackage{inconsolata}

\usepackage{graphicx}
\usepackage{booktabs}
\usepackage{subcaption}
\usepackage{subcaption}
\usepackage{amsmath}
\usepackage{amssymb}

\usepackage[most]{tcolorbox}

\definecolor{boxgray}{RGB}{246,247,249}

\tcbset{
  qualitative/.style={
    enhanced,
    breakable,
    colback=white,
    colframe=black!30,
    boxrule=0.4pt,
    arc=1pt,
    left=8pt,
    right=8pt,
    top=7pt,
    bottom=7pt,
    colbacktitle=boxgray,
    coltitle=black,
    fonttitle=\bfseries,
    toptitle=3pt,
    bottomtitle=3pt,
    before skip=8pt,
    after skip=8pt
  }
}

\newtcolorbox{questionbox}{
  qualitative,
  title={Problem}
}

\newtcolorbox{modelingbox}{
  qualitative,
  title={Model Output}
}

\title{Decoupled Physical Modeling and Execution for Physics Reasoning}

\author{
\textbf{Ye Zhang\textsuperscript{1}},
\textbf{Xuehang Guo\textsuperscript{2}},
\textbf{Rui Pan\textsuperscript{3}},
\textbf{Pengfei Yu\textsuperscript{4}},
\\
\textbf{Denghui Zhang\textsuperscript{5}},
\textbf{Manling Li\textsuperscript{6}},
\textbf{Qingyun Wang\textsuperscript{2}}
\\
\\
\textsuperscript{1}University of Pennsylvania,
\textsuperscript{2}William \& Mary,
\\
\textsuperscript{3}University of Illinois Urbana-Champaign,
\textsuperscript{4}Amazon,
\\
\textsuperscript{5}Stevens Institute of Technology,
\textsuperscript{6}Northwestern University
}

\begin{document}
\maketitle
\begin{abstract}
Physics reasoning requires constructing a consistent model of the underlying physical system rather than relying solely on symbolic or formula-based manipulation.
Although large language models have shown strong ability in solving math and coding problems, they still struggle with physics problems, as these problems entangle the physical modeling process with mathematical calculations.
Humans approach physics by first building a representation of the system before performing calculations. Inspired by this, we introduce a unified framework that distills intermediate representations that explicitly encode the physical modeling process and adopt a two-stage post-training strategy, where supervised fine-tuning establishes structured modeling, and reinforcement learning with rubric-based feedback improves the quality of the modeling process.
Experiments on multiple multimodal physics benchmarks show that our approach generally improves physical reasoning performance across different models and datasets. Across PhysReason, PhyX, and SeePhys, physical modeling outperforms GRPO by \verb|~|3\% on average. showing that explicit physical modeling is an effective strategy for improving physics reasoning in small VLMs.
\end{abstract}

\section{Introduction}
\label{introduction}

\begin{figure}[t]
    \centering
    \includegraphics[width=\columnwidth]{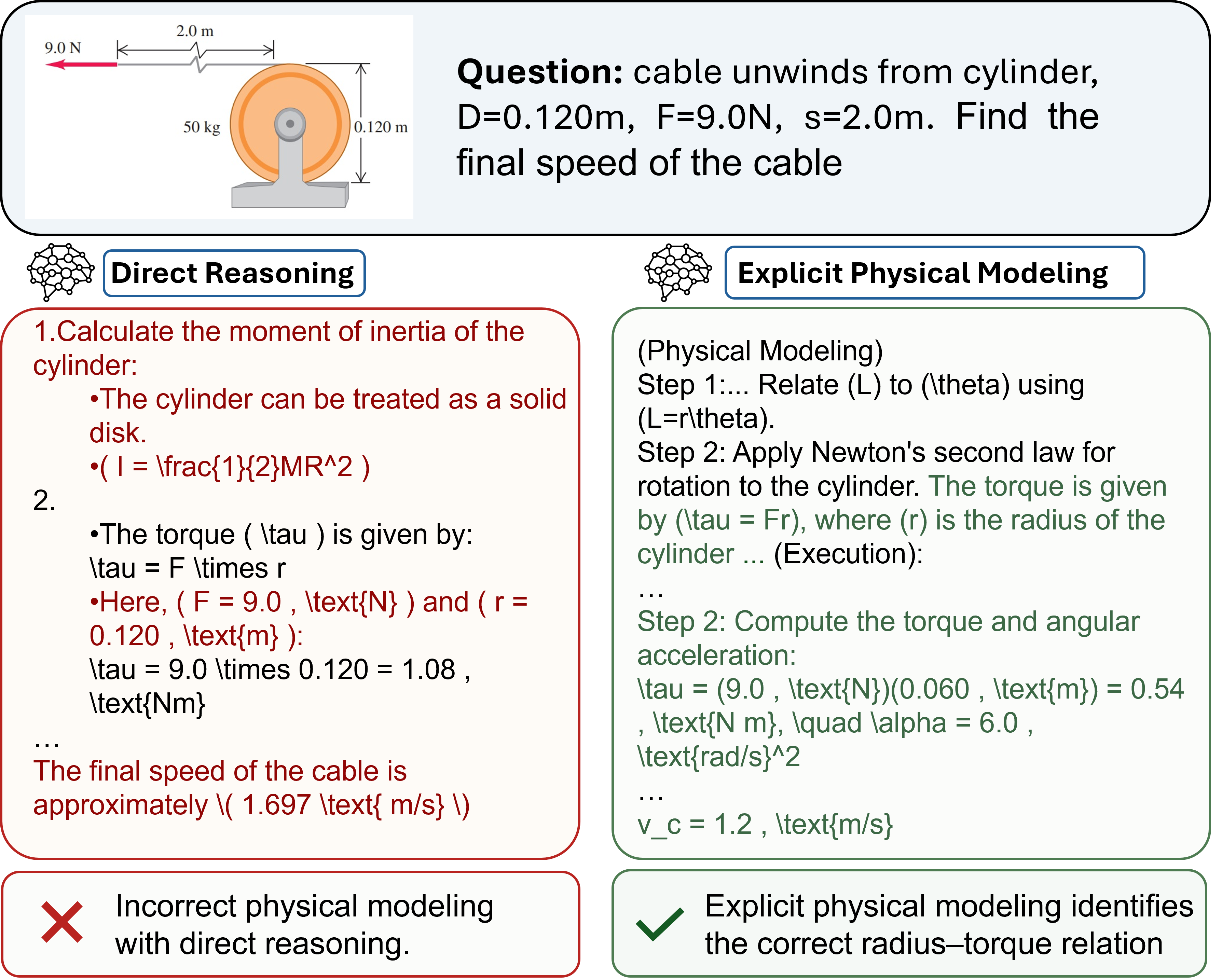}
    \caption{
    Physics reasoning errors often arise from incorrect physical modeling. Explicit modeling separates system representation from mathematical execution.
    }
    \label{fig:motivation}
\end{figure}

\begin{figure*}[t]
    \centering
    \includegraphics[width=\textwidth]{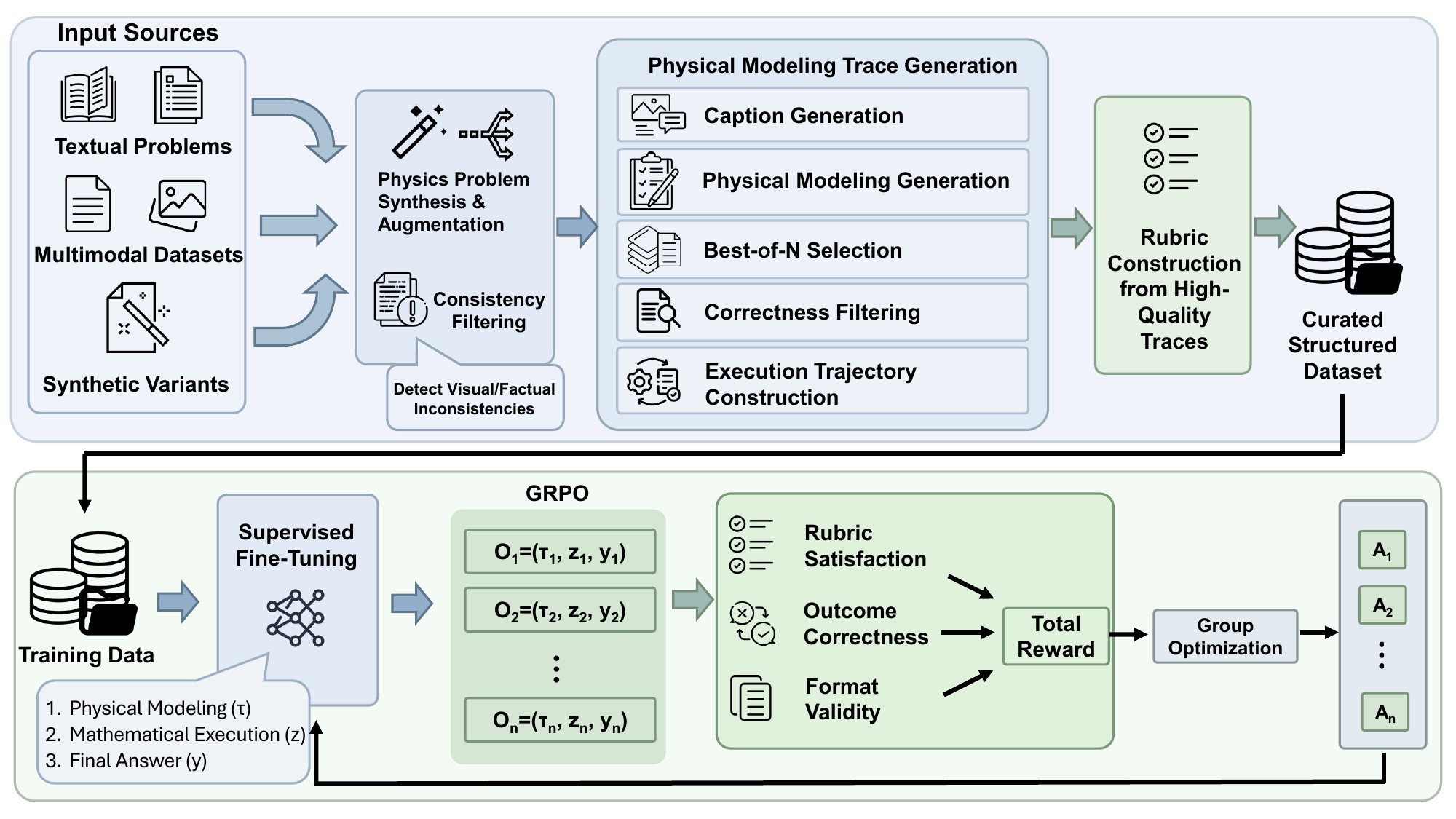}
    \caption{
    Overview of our physical-modeling post-training framework. Explicit physical modeling provides both structured SFT trajectories and rubric-based process rewards for GRPO, enabling the model to separate system modeling from mathematical execution.
    }
    \label{fig:main}
\end{figure*}

Solving physics problems requires more than selecting and applying equations. Human experts usually organize the problem into a structured physical representation that captures the relevant principles and relationships~\citep{RepresentationofPhysics, PhysRevSTPER.4.010111, ResearchonProblemSolvingPhysics}. In this view, physics reasoning involves constructing and operating on an internal model of the system, rather than performing symbolic manipulation alone.

Multimodal Large Language Models (MLLMs) have demonstrated strong performance in a wide range of reasoning tasks, particularly in mathematical~\citep{zhang2024mathverse,lu2024mathvistaevaluatingmathematicalreasoning,zhang2024mavismathematicalvisualinstruction,shi2024mathllavabootstrappingmathematicalreasoning,chen2025mint} and logical reasoning~\citep{xu2024largelanguagemodelsreally,xu2024faithfullogicalreasoningsymbolic,hendrycks2021measuringmassivemultitasklanguage,xu2026muslrmultimodalsymboliclogical}.
However, in physics reasoning, MLLMs tend to implicitly conflate physical modeling and mathematical execution into a single reasoning process, leading to conceptual errors in physics problem-solving. Our preliminary analysis of 500 responses further shows that a large fraction of errors involve incorrect or incomplete physical modeling (Appendix~\ref{appendix:preliminary_error_analysis}). In Figure~\ref{fig:motivation}, direct reasoning fails to identify the force arm by using the cylinder diameter radius relation incorrectly, leading to an overestimated angular acceleration and final cable speed.

Recent post-training methods improve reasoning by introducing an explicit plan before execution~\citep{jiao2024learningplanningbasedreasoningtrajectories,parmar2025plantuningposttraininglanguagemodels} or by scoring intermediate reasoning against multiple rubric criteria~\citep{gunjal2025rubricsrewardsreinforcementlearning,viswanathan2025checklistsbetterrewardmodels,zhou2025breaking,jayalath2026computeteacherturninginference}. Planning-based methods organize what the model should do at each stage, but do not directly check whether the underlying system has been represented correctly. Rubric-based methods provide more detailed feedback on intermediate reasoning, but their criteria often focus on the validity of individual steps rather than the physical model on which those steps are based. In physics, an incorrect representation of the system can make the entire derivation invalid, even when the subsequent steps appear reasonable. The reward should therefore assess the physics-specific constraints and principles that define the physical system. We provide a detailed discussion of related works in Appendix~\ref{appendix:related_work}.

In this paper, we explicitly model physics reasoning as a two-stage process consisting of \textbf{physical modeling} and \textbf{mathematical execution}. We introduce structured physical modeling traces as an intermediate representation and train models through a two-stage post-training framework. First, supervised fine-tuning (SFT) teaches the model to generate explicit physical modeling before execution. Then, Group Relative Policy Optimization (GRPO)~\cite{shao2024deepseekmathpushinglimitsmathematical} with rubric-guided rewards further improves the quality and physical grounding of the generated reasoning process. Experimental results show that explicitly modeling the physical system consistently improves performance on open-ended multimodal physics reasoning benchmarks.

\section{Structured Physical Modeling for Physics Reasoning}
\label{method}

\subsection{Overview and Task Formulation}
\label{method:overview}
We decompose the physics problem-solving into two stages: \textbf{physical modeling} and \textbf{mathematical execution}. In the \textbf{physical modeling} stage, the large language model applies relevant concepts and constraints to understand the system, while in the \textbf{mathematical execution} stage, these abstract physics concepts and constraints are formulated into specific equations for execution.
Given the input problem $x$, the model first generates $\tau$ to construct the physical modeling, and then produces $(z, y)$ based on this representation:
\begin{equation}
\pi_\theta(\tau, z, y \mid x) = \pi_\theta(z, y \mid \tau, x)\, \pi_\theta(\tau \mid x).
\end{equation}

To formalize structured reasoning, we conduct a two-stage post-training strategy. We first perform supervised fine-tuning (SFT) as a cold start.
We further refine this capability using Group relative Policy Optimization(GRPO)~\cite{shao2024deepseekmathpushinglimitsmathematical} with rubric-guided rewards, which improves modeling quality and encourages physically grounded and informative representations (see Figure~\ref{fig:main}).

\begin{table*}[t]
\centering
\small
\setlength{\tabcolsep}{2.5pt}
\begin{tabular}{lcccccccccc}
\toprule
\textbf{Method}
& \multicolumn{2}{c}{\textbf{PhysReason}}
& \multicolumn{2}{c}{\textbf{PhyX}}
& \multicolumn{2}{c}{\textbf{SeePhys}}
& \multicolumn{2}{c}{\textbf{PhysUni$_{\mathrm{mc}}$}}
& \multicolumn{2}{c}{\textbf{MMK-12$_{\mathrm{phys.}}$}} \\
\cmidrule(lr){2-3}
\cmidrule(lr){4-5}
\cmidrule(lr){6-7}
\cmidrule(lr){8-9}
\cmidrule(lr){10-11}
& \textbf{P@1} & \textbf{P@5}
& \textbf{P@1} & \textbf{P@5}
& \textbf{P@1} & \textbf{P@5}
& \textbf{P@1} & \textbf{P@5}
& \textbf{P@1} & \textbf{P@5} \\
\midrule

\multicolumn{11}{l}{\textbf{Qwen2.5-VL-3B-Instruct}} \\
Baseline 
& 13.3 & 33.6 
& 10.9 &  23.5
& 9.1 &  24.7
& 17.1 & 50.9 
& 35.0 & 73.4 \\

GRPO
& 17.3 &  36.8
& 12.6 &  27.0
& 12.4 &  29.8
& 19.1 &  52.7
& 42.4 &  82.4  \\ 

\textbf{Physics-Modeling + GRPO} 
& \textbf{22.9} & \textbf{47.1}
& \textbf{13.6} & \textbf{34.7}
& \textbf{16.5} & \textbf{33.9}
& \textbf{21.2} & \textbf{53.7}
& \textbf{46.2} & \textbf{84.0} \\ 
\midrule

\multicolumn{11}{l}{\textbf{Qwen2.5-VL-7B-Instruct}} \\
Baseline 
& 25.0 & 47.8 
& 17.0 & 35.7 
& 14.0 & 28.9
& 19.3 & 53.7 
& 45.4 & 79.4 \\

GRPO
& 28.7 & 52.5
& 19.2 & 39.3 
& 18.2 & 33.1
& 20.9 & 55.4 
& 54.2 & 87.4 \\

\textbf{Physics-Modeling + GRPO} 
& \textbf{32.1} & \textbf{58.2}
& \textbf{21.4} & \textbf{40.0}
& \textbf{19.8} & \textbf{34.7}
& \textbf{23.5} & \textbf{57.8}
& \textbf{56.8} & \textbf{89.6} \\
\midrule

\multicolumn{11}{l}{\textbf{Qwen3-VL-8B-Instruct}} \\
Baseline 
& 51.8 & 70.7
& 43.2 & 61.4 
& 27.3 & 40.5
& 32.4 & 60.2 
& 67.6 & 84.8 \\

GRPO
& \textbf{54.0} & 73.9 
& \textbf{46.1} & 64.1
& 30.6 & 45.5
& 34.9 & 63.4
& 73.8 &  90.0 \\

\textbf{Physics-Modeling + GRPO} 
& 53.3 & \textbf{74.7}
& 45.1 & \textbf{64.8}
& \textbf{32.2} & \textbf{47.9}
& \textbf{36.6} & \textbf{65.5} 
& \textbf{74.4} & \textbf{91.2} \\
\midrule

\multicolumn{11}{l}{\textbf{Gemma3-4B-IT}} \\
Baseline 
& 7.3 & 17.1
& 7.5 & 16.3 
& 5.0 & 16.3
& 17.8 & 48.9 
& 27.0 & 61.8 \\

GRPO
& 10.9 & 19.3 
& 9.5 & 17.0
& 9.9 & 18.2
& 21.6 & 49.9
& 32.6 & 64.0 \\

\textbf{Physics-Modeling + GRPO} 
& \textbf{13.1} & \textbf{22.7}
& \textbf{11.2} & \textbf{18.7}
& \textbf{11.6} & \textbf{19.0}
& \textbf{22.4} & \textbf{51.1} 
& \textbf{38.0} & \textbf{69.4} \\

\bottomrule
\end{tabular}
\caption{
Main results on PhysReason~\citep{zhang2025physreason}, PhyX~\citep{shen2025phyxdoesmodelwits}, and SeePhys~\citep{xiang2025seephys} for in-domain evaluation, and on PhysUniBench~\citep{wang2026physunibenchmultimodalphysicsreasoning} and the physics subset of MMK-12~\citep{meng2025mmeurekaexploringfrontiersmultimodal} for out-of-domain evaluation. We report Pass@1 and Pass@5 for the baseline and GRPO variants. Bold indicates the best result within each model block.
}
\label{tab:main_results}
\end{table*}

\subsection{Physics-Modeling Data Construction}
\label{method:data}

\paragraph{Structured Modeling Trace Generation.}
We construct the multimodal problem pool from PhysReason~\cite{zhang2025physreason}, PhyX~\cite{shen2025phyxdoesmodelwits}, and SeePhys~\cite{xiang2025seephys}, and the text-only pool from SCP-116K~\cite{lu2025scp116khighqualityproblemsolutiondataset} and MegaScience~\cite{fan2025megascience}.
We further synthesize additional multimodal problems that preserve the underlying physical principles and concepts by using Gemini-2.5-Flash~\cite{comanici2025gemini25pushingfrontier}. For each problem $x_i$, we generate a structured physical modeling trace $\tau_i$ by rejection sampling using Qwen3.5-Flash~\cite{qwen3.5}. To improve visual grounding, we use Qwen3-VL-32B-Instruct~\citep{bai2025qwen3vltechnicalreport} to extract image captions. We then apply a two-stage verification process to retain high-quality modeling traces. Finally, we obtain the execution trajectory $z_i$ by following the verified modeling trace, thereby deriving the correct solution. Details of the generation pipeline and dataset statistics are provided in Appendices~\ref{appendix:data} and~\ref{appendix:data_statistics}, respectively.

\paragraph{Rubric Derivation from Modeling Traces.}
We derive the rubrics from verified structured physical modeling traces. For each verified physical modeling trace, we generate a rubric set
$\mathcal{C}=\{c_k\}_{k=1}^{|\mathcal{C}|}$ with 8--12 criteria, covering the following aspects:
(1) \textbf{Visual Criteria};
(2) \textbf{Physical Modeling Criteria};
(3) \textbf{Strategic Planning Criteria};
(4) \textbf{Execution Criteria}.
Each criterion $c_k$ consists of a textual specification $d_k$ and an adaptive weight $w_k$, i.e., $c_k=(d_k,w_k)$. The weight reflects the criterion's relative importance for approaching the problem.
Details are provided in Appendix~\ref{appendix:rubric}.

\subsection{Two-Stage Post-Training}
\label{method:training}

We adopt a two-stage post-training pipeline. We perform SFT on structured trajectories $(\tau,z,y)$, where $\tau$ is the physical modeling, $z$ is the execution process, and $y$ is the final answer:
\begin{equation}
\mathcal{L}_{\text{SFT}} = - \mathbb{E}_{(x,\tau,z,y)\sim\mathcal{D}}
\left[\log \pi_\theta(\tau,z,y\mid x)\right].
\end{equation}
We use a text-to-multimodal curriculum: text-only data first establishes the modeling format, and multimodal data then teaches visual grounding.

We optimize the model with GRPO using a reward that combines final-answer correctness $R_{\text{outcome}}$, format validity $R_{\text{format}}$, and rubric-based evaluation $R_{\text{rubric}}$  of the intermediate physical modeling:
\begin{equation}
R_{\text{total}} =
R_{\text{outcome}} + R_{\text{format}} + R_{\text{rubric}}.
\end{equation}
Details are provided in the Appendix~\ref{appendix:grpo}.

\section{Results}
\label{experiment}

\paragraph{Performance of Physical Modeling.}
As shown in table~\ref{tab:main_results}, physical modeling consistently improves pass@1 accuracy over vanilla GRPO for both Qwen2.5-VL-Instruct~\cite{bai2025qwen25vltechnicalreport} backbones and Gemma3-4B-IT~\cite{gemmateam2025gemma3technicalreport}, while maintaining comparable performance on Qwen3-VL-8B-Instruct~\cite{bai2025qwen3vltechnicalreport}. The gains reach 5.6 points on PhysReason for Qwen2.5-VL-3B-Instruct and 5.4 points on MMK-12\textsubscript{phys} for Gemma3-4B-IT. This indicates that structured physical modeling traces may help elicit latent physical reasoning capability. On Qwen3-VL-8B-Instruct, vanilla GRPO may already elicit much of this capability, accounting for the marginal changes in pass@1.
The advantage of physical modeling becomes more consistent at pass@5. Physical modeling outperforms vanilla GRPO across all backbones and benchmarks, where the gains are substantial on Qwen2.5-VL-3B-Instruct, reaching 10.3 points on PhysReason and 7.7 points on PhyX. Improvements remain positive in Qwen3-VL-8B-Instruct, despite the marginal changes observed in pass@1. This suggests that physical modeling leads to a higher probability of obtaining a correct solution over multiple sampled responses.


\begin{table}[t]
\centering
\small
\setlength{\tabcolsep}{4pt}
\begin{tabular}{lcccccc}
\toprule
\textbf{Method}
& \multicolumn{2}{c}{\textbf{PhysReason}}
& \multicolumn{2}{c}{\textbf{PhyX}}
& \multicolumn{2}{c}{\textbf{SeePhys}} \\
\cmidrule(lr){2-3}
\cmidrule(lr){4-5}
\cmidrule(lr){6-7}
& \textbf{P@1} & \textbf{P@5}
& \textbf{P@1} & \textbf{P@5}
& \textbf{P@1} & \textbf{P@5} \\
\midrule
\multicolumn{7}{l}{\textbf{Qwen2.5-VL-7B}} \\
Baseline & 25.0 & 47.8 & 17.0 & 35.7 & 14.0 & 28.9 \\
SFT Cold Start & 27.2 & 54.4 & 18.2 & 36.0 & 15.7 & 28.9 \\
\hspace{1em}\textit{w/o Textual} & 22.4 & 43.8 & 15.1 & 29.9 &14.0 & 25.6 \\
\midrule
\multicolumn{7}{l}{\textbf{Qwen3-VL-8B}} \\
Baseline & 51.8 & 70.7 & 43.2 & 61.4 & 27.3 & 40.5 \\
SFT Cold Start & 50.3 & 70.4 & 41.5 & 61.8 & 29.8 & 42.1 \\
\bottomrule
\end{tabular}
\caption{Effect of Physical Modeling SFT Cold Start}
\label{tab:baseline_sft_cold_start}
\end{table}

\paragraph{Effect of SFT Cold Start.}
Table~\ref{tab:baseline_sft_cold_start} shows the effect of physical modeling of the SFT cold start on Qwen2.5-VL-7B-Instruct and Qwen3-VL-8B-Instruct. In Qwen2.5-VL-7B-Instruct, the proposed SFT cold start consistently improves performance on PhysReason, PhyX, and SeePhys (e.g., the Pass@1/Pass@5 scores on PhysReason increase from 25.0/47.8 to 27.2/54.4), whereas the changes in Qwen3-VL-8B-Instruct are relatively smaller and inconsistent, ranging from -1.7 to +2.5 points for Pass@1 and from -0.3 to +1.6 points for Pass@5. To test whether multimodal data alone are sufficient, we train Qwen2.5-VL-7B-Instruct without the initial textual SFT stage. Without the initial textual SFT stage, the performance of Qwen2.5-VL-7B-Instruct drops on all three benchmarks, especially PhysReason, where Pass@1/Pass@5 decreases from 27.2/54.4 to 22.4/43.8, indicating that multimodal SFT alone  does not account for the gains of the full curriculum. The initial textual physical modeling provides a useful foundation for subsequent multimodal training.

\begin{figure}[t]
    \centering
    \includegraphics[width=0.48\textwidth]{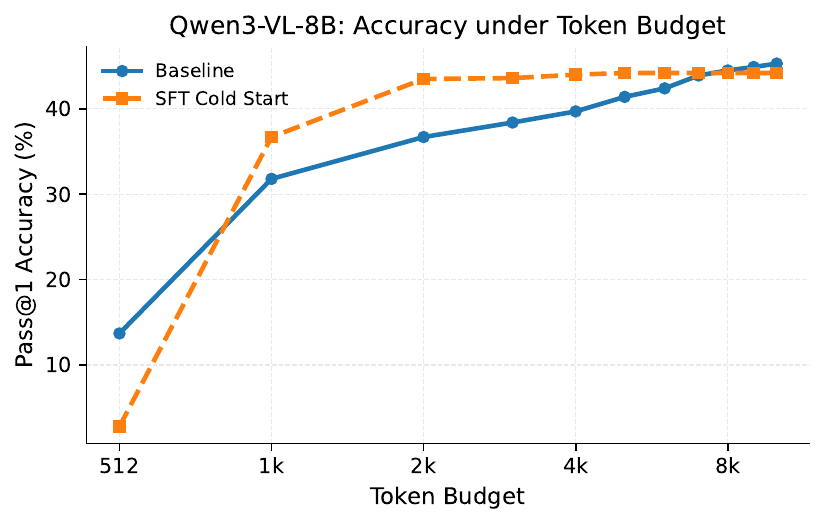}
    \caption{Pass@1 accuracy of Qwen3-VL-8B-Instruct under different token budgets.}
    \label{fig:qwen3_token_budget}
\end{figure}

\begin{table}[t]
\centering
\small
\setlength{\tabcolsep}{6pt}
\begin{tabular}{lcc}
\toprule
\textbf{Method} & \textbf{Samples w/ Reflection} & \textbf{Ratio} \\
\midrule
Baseline & 309/500 & 0.618 \\
Modeling-SFT & 5/500 & 0.010 \\
\bottomrule
\end{tabular}
\caption{Frequency of reflective reasoning patterns in 500 sampled Qwen3-VL-8B-Instruct generations.}
\label{tab:reflection_behavior}
\end{table}

\paragraph{Token Efficiency in Qwen3-VL-8B}
To better understand how the SFT cold start affects Qwen3-VL-8B-Instruct beyond the final accuracy, we analyze the model under different token budgets and examine its behavior of reflections. At 512 tokens, the SFT model performs below the baseline. It outperforms the baseline at 1k tokens, while increasing the token budget beyond 2k tokens gains little further improvement (See Figure~\ref{fig:qwen3_token_budget}). The baseline continues to improve as the token budget increases and approaches the SFT cold-start model only at larger token budgets. We further quantify explicit reflection by counting responses that contain reflective expressions such as "wait", "rethink", and "verify". These reflective keywords indicate the reflective patterns within model outputs. We sample 500 responses from Qwen3-VL-8B-Instruct and the SFT cold-start model on the same questions. The reflection rate drops from 61.8\% in the baseline to 1.0\% after SFT cold start among the 500 sampled responses, as reported in Table~\ref{tab:reflection_behavior}. Qwen3-VL-8B-Instruct appears to spend additional tokens revisiting and correcting earlier steps, whereas after SFT cold start, the model is more likely to establish the correct formulation before calculation and carry out the subsequent derivation without repeated correction. This accounts for its better performance under smaller token budgets, even though the baseline catches up when longer responses are allowed. Qualitative examples are provided in Appendix~\ref{appendix:reflective_reasoning}.

\paragraph{Ablation Studies.}
Table~\ref{tab:ablation} presents the ablation analysis of our two main components. 
Removing the SFT cold start lowers the average Pass@1 score by 3.7 points, with a drop on every benchmark. Removing the rubric reward has a smaller effect, lowering the average by 1.3 points. The full method improves over this variant on four of the five benchmarks, with the largest gain of 2.8 points on MMK-12, while the two methods tie on SeePhys. The two components play different roles: SFT teaches the model to produce structured physical modeling traces, and the rubric provides feedback on their quality during GRPO. With both components, the model achieves the best or tied-best result on every benchmark.

\begin{table}[t]
\centering
\small
\setlength{\tabcolsep}{2.5pt}
\begin{tabular}{lccccc}
\toprule
\textbf{Method} 
& \textbf{PR} 
& \textbf{PhyX} 
& \textbf{SP} 
& \textbf{PUB} 
& \textbf{MMK} \\
\midrule
w/o SFT 
& 28.1 & 18.2 & 15.7 & 20.5 & 52.8 \\
w/o Rubric 
& 30.4 & 20.9 & 19.8 & 21.9 & 54.0 \\
\textbf{Full} 
& \textbf{32.1} & \textbf{21.4} & \textbf{19.8} & \textbf{23.5} & \textbf{56.8} \\
\bottomrule
\end{tabular}
\caption{Component ablations on Qwen2.5-VL-7B-Instruct. We report Pass@1 accuracy on PhysReason (PR), PhyX, SeePhys (SP), PhysUniBench (PUB), and MMK-12 (MMK).}
\label{tab:ablation}
\end{table}

\begin{figure}[t]
    \centering
    \includegraphics[width=0.4\textwidth]{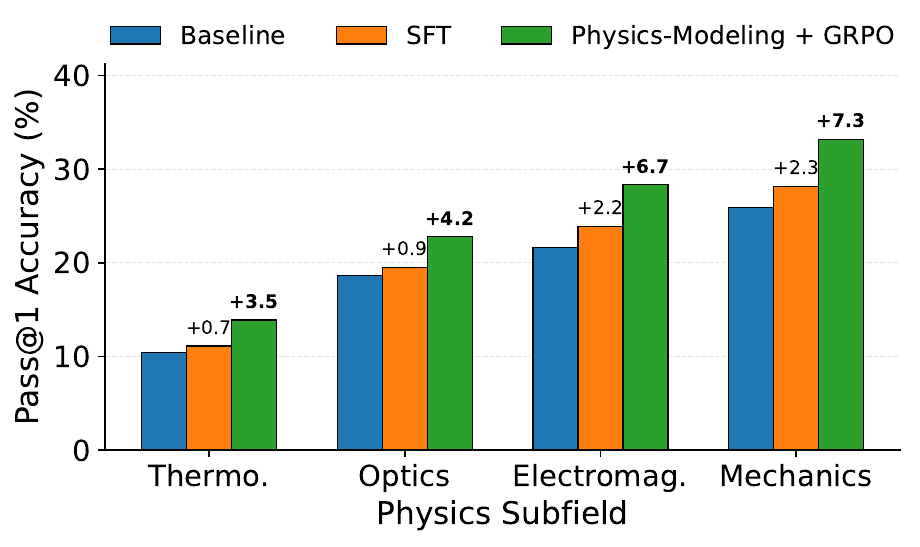}
    \caption{Performance across physics subfields for Qwen2.5-VL-7B-Instruct.}
    \label{fig:subfield_analysis}
\end{figure}

\paragraph{Subfield-wise Analysis.}
We further analyze performance across mechanics, electrodynamics, optics, and thermodynamics. Figure~\ref{fig:subfield_analysis} reports the accuracies for the four physics subfields. The gains are largest in mechanics and electrodynamics, at 7.3 and 6.7 points over the baseline. Optics and thermodynamics improve by 4.2 and 3.5 points. Many mechanics and electrodynamics problems depend on first identifying the relevant objects and then specifying the forces, fields, and constraints between them. The improvements are smaller for optics and thermodynamics. Optics problems are often built around ray diagrams, whereas thermodynamics problems involve state changes and energy flow.


\paragraph{Qualitative Analysis.}
The Carnot engine-refrigerator example (see Appendix~\ref{Appendix: Qualitative_Analysis}) illustrates how structured modeling supports reasoning in coupled physical systems. The model first analyzes the heat engine to determine the available work output, and then transfers this quantity to the refrigerator stage, enabling a coherent multi-step solution.
This example highlights that the benefit of structured physical modeling lies in explicitly organizing key variables, constraints, and governing relations before numerical execution. Such intermediate representations provide a clear pathway that guides subsequent reasoning steps.
We observe similar patterns across other subfields.

\section{Conclusion}
In this paper, we study physics reasoning from the perspective of structured physical modeling. We formulate physics reasoning as a two-stage process that separates physical modeling from mathematical execution, and introduce an intermediate representation to explicitly capture the modeling stage. To support this formulation, we develop a data construction pipeline for structured supervision and train the model through supervised fine-tuning followed by rubric-based reinforcement learning. The physics-specific rubric signals are derived from high-quality modeling traces and provide fine-grained feedback on whether the constructed physical model is consistent and complete. Experiments across multiple multimodal physics benchmarks show that explicitly improving physical modeling leads to better reasoning performance. The improvements are consistent at Pass@5 across all evaluated models, while the Pass@1 changes on Qwen3-VL-8B-Instruct are relatively small. Our token-budget analysis further shows that the SFT cold start allows Qwen3-VL-8B-Instruct to maintain strong performance with shorter reasoning traces and less explicit reflection.

\section*{Limitations}
Our study shows that incorporating explicit physical modeling into post-training improves the physics reasoning capabilities of LLMs. However, our method relies on an automated pipeline to generate physical modeling traces and physics-specific rubrics, whose quality depends on the reasoning capabilities of the models that generate them. Although a multi-stage filtering process removes many erroneous examples, the resulting data may still contain incorrect assumptions or incomplete constraints, and the generated rubrics may not cover alternative but valid formulations. The extent of the improvements varies across model backbones, as the Pass@1 results for Qwen3-VL-8B-Instruct are mixed. Moreover, the token-budget and reflection analyses reveal changes in reasoning behavior after SFT, although the connection between these changes and token efficiency requires further investigation. The experiments described here focus on multimodal physics reasoning and therefore do not examine generalization to other scientific fields. Finally, rubric-based RL incurs additional training costs because each sampled response must be evaluated by LLM-as-a-judge.

\section*{Acknowledgments}
The authors acknowledge William \& Mary Research Computing for providing computational resources and/or technical support that have contributed to the results reported within this paper. This work used DeltaAI at NCSA through allocation CIS260012 and CIS230280 from the Advanced Cyberinfrastructure Coordination Ecosystem: Services \& Support (ACCESS) program, which is supported by U.S. National Science Foundation grants \#2138259, \#2138286, \#2138307, \#2137603, and \#2138296. This material is based upon work supported by the Google Cloud Research Credits program with the award GCP19980904.

\bibliography{custom}

\appendix

\section{Related Work}
\label{appendix:related_work}

\paragraph{LLMs for Physics Reasoning.}
Recent work has introduced a variety of benchmarks for evaluating physics reasoning across both textual and multimodal settings, covering a broad range of difficulty levels from high-school to advanced problems~\citep{wang2026physunibenchmultimodalphysicsreasoning,zhang2025physreason,feng2025physicsbenchmarkingfoundationmodels,xiang2025seephys}. 
Early works treat physics reasoning as part of general scientific benchmarks~\citep{lu2022learnexplainmultimodalreasoning,wang2024mmluprorobustchallengingmultitask,rein2023gpqagraduatelevelgoogleproofqa}, while more recent work focuses on dedicated physics benchmarks that emphasize more challenging and diverse reasoning scenarios. In particular, multimodal physics reasoning requires models to jointly reason over textual and visual information to construct and solve physically grounded representations of the underlying system~\citep{dai2025physicsarenamultimodalphysicsreasoning,shen2025phyxdoesmodelwits,zhang2025physreason,xiang2025seephys}.
Across these benchmarks, physics reasoning is typically characterized by a modeling-centric process, where the large language model must construct a structured representation of the physical system before performing symbolic or numerical computation~\citep{dai2025physicsarenamultimodalphysicsreasoning}.
Prior works have explored approaches such as knowledge-augmented reasoning~\citep{pang2024physicsreasonerknowledgeaugmentedreasoning} or iterative refinement methods~\citep{jaiswal2024improvingphysicsreasoninglarge} that aim to enhance intermediate reasoning quality. A recent study further suggests that scaling inference-time computation alone is often insufficient and that effective physics reasoning is typically based on more structured and consistent reasoning patterns~\citep{gao2025testtimescalingtechniquestheoretical,dan2025symbolicnumericalunderstandingphysics}. These observations highlight the importance of explicitly modeling the underlying physical system, rather than relying on implicit generation without structured modeling supervision.

\paragraph{Post-Training Methods for Reasoning.}
Reinforcement learning with verifiable rewards (RLVR) has shown strong performance in domains such as mathematics and coding by leveraging programmatic or externally verifiable rewards~\citep{shao2024deepseekmathpushinglimitsmathematical}.
Beyond outcome-level rewards, prior work explores richer feedback over reasoning trajectories~\citep{cui2025processreinforcementimplicitrewards,lambert2025tulu3pushingfrontiers}. Process supervision and planning-style training provides denser guidance over intermediate reasoning steps by encouraging structured decomposition of the solution process~\citep{jiao2024learningplanningbasedreasoningtrajectories,parmar2025plantuningposttraininglanguagemodels}.
More recently, rubric- or checklist-based evaluation has emerged as a way to assess open-ended reasoning tasks using multiple explicit criteria rather than relying solely on final-answer accuracy or holistic judgments~\citep{arora2025healthbenchevaluatinglargelanguage, galvansosa2025rubrikscubetestingnew}. Beyond evaluation, such structured criteria are used as rubric-based rewards to evaluate the reasoning trajectory during policy optimization in reinforcement learning~\citep{gunjal2025rubricsrewardsreinforcementlearning, viswanathan2025checklistsbetterrewardmodels,zhou2025breaking, jayalath2026computeteacherturninginference}.
Despite providing richer supervision than final-answer rewards alone, these approaches largely rely on generic criteria for assessing reasoning quality. In physics, however, effective problem-solving depends on constructing an explicit physical representation of the system, thereby motivating supervision signals grounded in the underlying physical structure.

\section{Preliminary Analysis}
\label{appendix: preliminary_analysis}
To identify where the main difficulties in physics reasoning arise, we conduct two preliminary analyses. We first examine the distribution of errors in generated solutions and then keep the executor fixed while varying the physical modeling trace provided to it.

\subsection{Error Analysis}
\label{appendix:preliminary_error_analysis}
\begin{figure}
    \centering
    \includegraphics[width=1.0\linewidth]{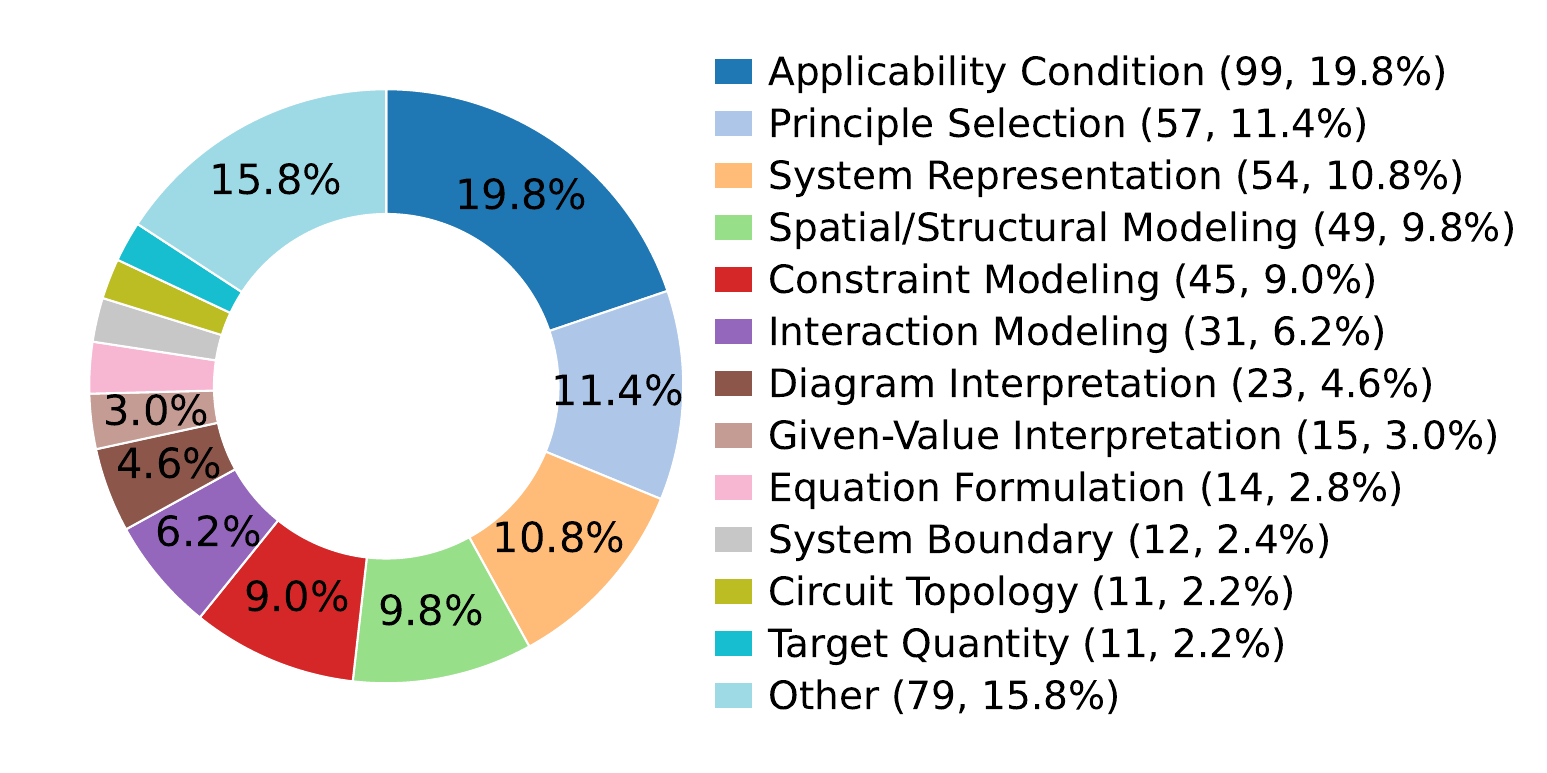}
    \caption{
Distribution of physics reasoning error types from a preliminary analysis using a Qwen2.5-VL-7B-Instruct.}
    \label{fig:preliminary_error_analysis}
\end{figure}

To better understand the challenges of physics reasoning, we conduct a preliminary error analysis using Qwen2.5-VL-7B-Instruct~\cite{bai2025qwen25vltechnicalreport} on a subset of 500 physics problems. We categorize the incorrect responses into different error types using Qwen3.5-Flash~\cite{qwen3.5}. As shown in Figure~\ref{fig:preliminary_error_analysis}, a large fraction of failures arises from incorrect or incomplete physical modeling. This observation motivates our focus on explicitly separating physical modeling from downstream computation.

\begin{table}[t]
\centering
\small
\begin{tabular}{lcc}
\toprule
\textbf{Modeling Trace} & \textbf{Correct} & \textbf{Accuracy} \\
\midrule
No modeling trace     & 72/400  & 18.0\% \\
Self-generated trace  & 80/400  & 20.0\% \\
Strong-model trace    & 154/400 & 38.5\% \\
\bottomrule
\end{tabular}
\caption{Performance of a fixed executor under different sources of physical modeling traces on the same 400 problems.}
\label{tab:trace_quality}
\end{table}

\subsection{Modeling Trace Intervention}
\label{appendix:modeling_trace_intervention}
To understand how the modeling trace quality affects mathematical execution in physical reasoning, we conduct a fixed-executor diagnosis on 400 examples. The goal is to change the quality of the physical-modeling input while keeping the mathematical executor unchanged. We use Qwen2.5-VL-7B-Instruct~\cite{bai2025qwen25vltechnicalreport} as the same weak executor to solve each problem under three inputs: no trace, its own modeling trace, and a stronger-model modeling trace from Qwen3.5-Flash~\cite{qwen3.5}. The results are reported in Table~\ref{tab:trace_quality}.
Since the executor is unchanged, the main variable is the quality of the physical-modeling input. The gain from 80/400 to 154/400 suggests that better physical modeling can substantially improve final solving accuracy. This provides a more direct diagnostic of the mechanism underlying the method.

\section{Data Generation and Curation Details}
\label{appendix:data}

\subsection{Data Source}
\label{appendix:data_source}

\textbf{PhysReason}~\citep{zhang2025physreason} is a multimodal physics reasoning benchmark constructed from diverse sources including college entrance exams and international physics competitions. It consists of 1,200 carefully curated problems with diagram-grounded contexts, structured solutions, and step-level annotations, emphasizing long-horizon, multi-step reasoning. Notably, each problem often contains multiple sub-questions, which we decompose into independent samples to construct finer-grained training instances.

\textbf{PhyX}~\citep{shen2025phyxdoesmodelwits} focuses on visual physics reasoning across diverse scenarios. It contains 3,000 curated questions spanning six reasoning types, 25 sub-domains, and six core physics domains, including mechanics, electromagnetism, thermodynamics, optics, modern physics, and wave-related topics. Unlike purely symbolic benchmarks, PhyX emphasizes integrating visual understanding with physical principles.

\textbf{SeePhys}~\citep{xiang2025seephys} emphasizes vision-dependent physics reasoning grounded in diagrams. It contains 2,000 curated questions paired with 2,245 images, covering 7 physics domains and 21 heterogeneous diagram categories across knowledge levels from middle school to PhD qualifying exams. A substantial portion of the benchmark is vision-essential, requiring models to extract indispensable information directly from diagrams rather than relying solely on textual cues.

\textbf{SCP-116K}~\citep{lu2025scp116khighqualityproblemsolutiondataset} is a large-scale dataset of high-quality problem–solution pairs designed for scientific reasoning across STEM domains. It contains 116,756 rigorously curated examples automatically extracted from heterogeneous educational resources, including textbooks and academic materials. The dataset is constructed via a generalized multi-stage pipeline with stringent filtering to ensure scientific rigor and appropriate educational levels. Unlike manually curated benchmarks, SCP-116K emphasizes scalable data construction, providing both the dataset and the extraction pipeline to facilitate future extensions and domain transfer.

\textbf{MegaScience}~\cite{fan2025megascience} is a large-scale scientific reasoning dataset consisting of 1.25 million instances of diverse STEM domains. It is designed to support complex multi-step reasoning and scientific problem solving.

We construct our training data from both textual and multimodal physics sources. For textual data, we sample 12,000 physics-related problem–solution pairs from SCP-116K and MegaScience. For multimodal data, we randomly sample instances from PhysReason, PhyX, and SeePhys.
We first filter out non-English problems to ensure consistency. To establish a clean evaluation protocol, we reserve a held-out set of 1,000 problems, including 467 from PhysReason, 412 from PhyX, and 121 from SeePhys. These samples are excluded from all stages of data generation and training.
The remaining data is used as input to our data generation pipeline for constructing structured physical modeling traces and rubrics .

\subsection{Data Generation Pipeline}
\label{appendix:data_generation_pipeline}

\paragraph{Problem Augmentation.}
To expand the multimodal physics problem pool, we use Gemini-2.5-Flash~\citep{comanici2025gemini25pushingfrontier} to synthesize additional problem instances from existing multimodal physics data. Specifically, we apply the following transformations:
\begin{itemize}
    \item \textbf{Objective variation:} modifying the target quantity or question objective while keeping the physical setup unchanged;
    \item \textbf{Numerical perturbation:} altering numerical values (e.g., masses, distances, or constants) within physically valid ranges;
    \item \textbf{Linguistic rephrasing:} rewriting problem descriptions to introduce surface-level diversity without changing semantics.
\end{itemize}
We further filter out samples with visual conflicts or underlying physics inconsistencies using Gemini-2.5-Flash~\cite{comanici2025gemini25pushingfrontier}.

\textbf{Visual Captioning.}
To improve visual grounding, we first extract a caption from the image using Qwen3-VL-32B-Instruct~\cite{bai2025qwen3vltechnicalreport}. The caption is constrained to describe only observable and physically relevant elements in the scene, including object identities, spatial relationships, geometric configurations, and annotated quantities (e.g., angles, lengths, or directions), without introducing additional assumptions or inferred conclusions.

\textbf{Structured Physical Modeling.}
We construct the model input by combining the image, question, generated caption, and reference solution. The reference solution is used only to guide the modeling structure and does not directly determine the final answer during modeling trace generation.
We adopt a best-of-$N$ strategy ($N=5$) to generate multiple candidate physical modeling traces. Each trajectory is required to explicitly represent the physical system in a structured form, rather than a high-level summary. Specifically, each trajectory includes:
\begin{itemize}
    \item \textbf{System Understanding:} the target physical quantity or system property to be solved;
    \item \textbf{Constraints:} conditions implied by the problem setup, diagram, or geometry, such as boundary conditions, conservation constraints, or kinematic relations;
    \item \textbf{Governing principles:} the physical laws or principles relevant to the problem, along with how they apply to the current system;
    \item \textbf{Equations and physical relationships:} symbolic formulations that instantiate the governing principles and explicitly connect variables through the underlying physical relationships.
\end{itemize}

\paragraph{Two-stage Quality Control.}
To ensure that the generated modeling traces provide reliable supervision for physical modeling, we apply a two-stage filtering procedure at both the modeling process and outcome levels.

\textbf{Stage 1: Modeling-level filtering.}
We first evaluate each candidate modeling trace using Gemini-2.5-Flash, focusing on three aspects:
(1) \emph{internal coherence}, which checks whether the reasoning steps are logically consistent and free of contradictions;
(2) \emph{multimodal consistency}, which verifies that the modeling process is grounded in the visual scene and does not conflict with the image or problem description;
and (3) \emph{modeling specificity}, which requires the modeling trace to explicitly represent key elements of the physical system, including variables, constraints, governing principles, and their relationships.
Traces that are inconsistent, underspecified, or remain at a generic high-level description are discarded.

\textbf{Stage 2: Execution-based validation.}
Among the physical modeling traces that pass the first stage, we further validate whether the modeling trace can support correct downstream reasoning.
Specifically, we provide the question, image, and candidate modeling trace to Gemini-2.0-Flash\citep{geminiteam2025geminifamilyhighlycapable} and require the model to strictly follow the trace to derive the final answer.
Only traces that lead to correct solutions are retained.

This two-stage procedure ensures that the retained modeling traces are both physically grounded at the modeling stage and executable toward correct final answers.

\subsection{Rubric Construction}
\label{appendix:rubric}

\paragraph{Rubric Derivation from Physical Modeling Traces.}
For each problem, we derive a set of fine-grained evaluation criteria from the filtered modeling trace. Each criterion corresponds to a specific requirement of the modeling process, ensuring that the rubric is grounded in the underlying physical representation rather than generic reasoning patterns.

\paragraph{Design Principles.}
To ensure that the rubrics provide meaningful and reliable supervision, we enforce that each criterion is \emph{atomic}, capturing a single aspect of the modeling process; \emph{verifiable}, corresponding to a concrete and independently checkable requirement; and \emph{physically grounded}, reflecting explicit physical concepts such as variables, constraints, governing principles, and their relationships. In addition, we explicitly avoid vague or generic criteria (e.g., ``correct reasoning'' or ``proper simplification'') to ensure that all criteria are non-trivial and informative.

\paragraph{Rubric Structure.}
Each rubric consists of 8-12 criteria that collectively capture the construction of a physical system representation. Specifically, the criteria cover key aspects of the modeling process, including:
\begin{itemize}
    \item \textbf{Visual Criteria:} Verify the correctness of key information from physics diagrams (e.g., numerical values and spatial relationships);
    \item \textbf{Physical Modeling Criteria:} Evaluate the construction of the underlying physical modeling (e.g., concepts, constraints, and physics process);
    \item \textbf{Strategic Planning Criteria:} Evaluate the decomposition of a complex physics problem into appropriate subproblems;
    \item \textbf{Execution Criteria:} Evaluate the translation from abstract concepts to equations and symbolic representations, and intermediate answers.
\end{itemize}

This process converts modeling traces into structured, physically grounded rubrics that provide explicit supervision over the modeling process.

\subsection{Validation of Rubric-Based Judgments}
\label{appendix:rubric_validation}

We manually audit 50 randomly sampled training instances, covering 498 criterion-level judgments. Two human annotators independently evaluate whether each response satisfies the corresponding rubric criterion and achieve a raw agreement rate of 92.0\%. Disagreements are resolved through discussion to obtain consensus annotations. Using these annotations as the human reference, the LLM judge agrees on 439 of the 498 judgments, corresponding to an agreement rate of 88.2\%. The disagreements include 37 false positives and 22 false negatives.

\section{Data Statistics}
\label{appendix:data_statistics}
In Table~\ref{tab:appendix_data_statistics}, we provide detailed statistics of the constructed multimodal physical modeling trace dataset. The final dataset contains 8,880 physical trajectories, including 4,852 original samples and 4,028 synthesized samples, with relatively balanced coverage across major physics categories.
\begin{table}[t]
\centering
\scriptsize
\setlength{\tabcolsep}{3.5pt}
\begin{tabular}{llrrr}
\toprule
\textbf{Data} & \textbf{Cat.} & \textbf{Orig.} & \textbf{Synth.} & \textbf{Total} \\
\midrule
PhysReason & Electro. & 962 & 523 & 1485 \\
& Mech. & 567 & 476 & 1043 \\
& Optics & 353 & 207 & 560 \\
& Thermo. & 239 & 89 & 328 \\
& Other & 19 & 24 & 43 \\
\midrule
PhyX & Electro. & 418 & 454 & 872 \\
& Mech. & 644 & 695 & 1339 \\
& Optics & 537 & 469 & 1006 \\
& Thermo. & 367 & 348 & 715 \\
& Other & 133 & 110 & 243 \\
\midrule
SeePhys & Electro. & 227 & 249 & 476 \\
& Mech. & 212 & 254 & 466 \\
& Optics & 112 & 102 & 214 \\
& Thermo. & 43 & 25 & 68 \\
& Other & 19 & 3 & 22 \\
\midrule
\textbf{Total} & -- & \textbf{4852} & \textbf{4028} & \textbf{8880} \\
\bottomrule
\end{tabular}
\caption{Category-level statistics of the constructed physical modeling trace dataset.}
\label{tab:appendix_data_statistics}
\end{table}

\begin{table}[t]
\centering
\small
\setlength{\tabcolsep}{4pt}
\resizebox{\columnwidth}{!}{
\begin{tabular}{lrrrrr}
\toprule
& & \multicolumn{2}{c}{Similarity $>0.90$}
& \multicolumn{2}{c}{Near-Duplicate} \\
\cmidrule(lr){3-4}
\cmidrule(lr){5-6}
\textbf{Dataset} & \textbf{Size}
& \textbf{Count} & \textbf{Rate}
& \textbf{Count} & \textbf{Rate} \\
\midrule
PhysReason & 467  & 12 & 2.57\% & 6  & 1.28\% \\
PhyX       & 412  & 6  & 1.46\% & 4  & 0.97\% \\
SeePhys    & 121  & 5  & 4.13\% & 0  & 0.00\% \\
\midrule
Total      & 1000 & 23 & 2.30\% & 10 & 1.00\% \\
\bottomrule
\end{tabular}
}
\caption{Textual overlap between the training and held-out evaluation sets} 
\label{tab:data_overlap}
\end{table}

The PhysReason, PhyX, and SeePhys evaluation sets reported in Table~\ref{tab:main_results} are the 1,000 held-out problems described in Appendix~\ref{appendix:data}, which are excluded from every stage of data construction and augmentation. Synthetic problems are generated only from the remaining training pool, and no evaluation sample is used as a source for augmentation. 

We conduct an explicit overlap analysis between the training and evaluation sets using cosine similarity over semantic embeddings obtained from \texttt{sentence-transformers/all-MiniLM-L6-v2}. We manually inspect all 23 high-similarity cases. We define near-duplicates as problems sharing a similar physical setup and target quantity, with differences mainly in wording and numerical values. Only 10 evaluation problems have near-duplicate counterparts in the training set under this criterion (Figure~\ref{tab:data_overlap}).
We note that high embedding similarity does not necessarily indicate duplication. For physics problems, two instances can have similar physical scenarios but ask for different target quantities. Furthermore, even small changes in numerical values would change the implicit boundary conditions or constraints, although other physical settings are very close. Therefore, solving them may require different constraints or reasoning paths.

\section{Two-stage Post-training for Physical Modeling}
\label{appendix:grpo}
\subsection{SFT Cold Start for Structured Physical Modeling}

We first fine-tune the model to generate structured reasoning trajectories $(\tau, z, y)$, where $\tau$ denotes physical modeling, $z$ denotes execution, and $y$ denotes the final answer:
\begin{equation}
\mathcal{L}_{\text{SFT}} = - \mathbb{E}_{(x, \tau, z, y) \sim \mathcal{D}} \left[ \log \pi_\theta(\tau, z, y \mid x) \right].
\end{equation}
We use a curriculum from text-only to multi-modal data. The former establishes the structural pattern of modeling, while the latter teaches the models to ground physical modeling in visual information.

\subsection{Physical Modeling-Guided Policy Optimization}

We further optimize the model with rubric-based GRPO. 
Unlike standard GRPO, which primarily rewards final-answer correctness, our approach additionally rewards intermediate physical modeling and mathematical execution.
Specifically, given a group of sampled responses $\{y_i\}_{i=1}^{G}$ for each problem $x$, we optimize
\[
\mathcal{J}_{\mathrm{GRPO}}(\theta)
=
\mathbb{E}_{x,\{y_i\}}
\left[
\frac{1}{G}\sum_{i=1}^{G}
\hat{A}_i
\log \pi_\theta(y_i \mid x)
\right],
\]
where $\hat{A}_i$ is computed from the normalized reward within the sampled group. 
The reward combines final-answer correctness with rubric-based scores for physical modeling and execution.
\paragraph{Rubric Reward.}
Given a sampled reasoning trajectory $(\tau,z,y)_i$ and a set of fine-grained rubrics
$\mathcal{C}=\{(d_k,w_k)\}_{k=1}^{|\mathcal{C}|}$, we evaluate each criterion independently with an LLM-as-a-judge. 
The judge evaluates whether $(\tau,z,y)_i$ satisfies the criterion description $d_k$ under the original question $q$ and returns a binary score:
\begin{equation}
s_k(q,(\tau,z,y)_i)=
\begin{cases}
1, & \text{criterion satisfied}, \\
0, & \text{otherwise}.
\end{cases}
\end{equation}
The rubric reward is computed as the weighted average of criterion-level scores:
\begin{equation}
R_{\text{rubric}}(q,(\tau,z,y)_i)
=
\frac{\sum_{k=1}^{|\mathcal{C}|} w_k s_k(q,(\tau,z,y)_i)}
{\sum_{k=1}^{|\mathcal{C}|} w_k}.
\end{equation}

\paragraph{Outcome and Format Rewards.}
We use an outcome reward to evaluate final-answer correctness:
\begin{equation}
R_{\text{outcome}} =
\begin{cases}
1, & \hat{y}=y, \\
0, & \text{otherwise}.
\end{cases}
\end{equation}
We also use a format reward $R_{\text{format}}$, set to $0.5$ if the output follows the required \texttt{<plan>}, \texttt{<solution>}, and \texttt{<answer>} structure, and $0$ otherwise:
\begin{equation}
R_{\text{format}} =
\begin{cases}
0.5, & \text{valid format}, \\
0, & \text{otherwise}.
\end{cases}
\end{equation}

\paragraph{Total Reward.}
The total reward combines rubric, outcome, and format rewards:
\begin{equation}
R_{\text{total}}
=
R_{\text{rubric}} + R_{\text{outcome}} + R_{\text{format}}.
\end{equation}

\section{Experimental Setup}
\label{appendix:setup}

\paragraph{Datasets.}
For text-only SFT, we construct a 12K-problem training set based on SCP-116K~\citep{lu2025scp116khighqualityproblemsolutiondataset} and MegaScience~\citep{fan2025megascience}. 
For multimodal SFT, we use PhysReason~\citep{zhang2025physreason}, PhyX~\citep{shen2025phyxdoesmodelwits}, and SeePhys~\citep{xiang2025seephys} as the problem pool, holding out 1,000 problems for evaluation. We synthesize new problems from the remaining pool using Gemini-2.5-Flash~\cite{comanici2025gemini25pushingfrontier}, resulting in 8,880 multimodal training instances after synthesis and filtering. For GRPO, we further derive fine-grained rubrics from the verified structured modeling traces in the multimodal training set, enabling process-level reward of physical modeling.

\paragraph{Base Models.}
To evaluate our method, we fine-tune three open-source vision-language models of different scales and series: \textbf{Qwen2.5-VL-3B-Instruct}~\citep{bai2025qwen25vltechnicalreport}, \textbf{Qwen2.5-VL-7B-Instruct}~\citep{bai2025qwen25vltechnicalreport}, \textbf{Qwen3-VL-8B-Instruct}~\citep{bai2025qwen3vltechnicalreport}, and \textbf{Gemma3-4B-IT}~\citep{gemmateam2025gemma3technicalreport}

\paragraph{Implementation Details.}
For SFT, we adopt a curriculum training pipeline. We first perform text-only supervised fine-tuning for 2 epochs, followed by multimodal fine-tuning for 2 epochs on the constructed physics training set. We use LoRA~\citep{hu2021loralowrankadaptationlarge} with rank 64 and $\alpha=128$, and set the learning rate to $1\times10^{-4}$. SFT is implemented with \texttt{LLaMA-Factory}~\citep{zheng2024llamafactoryunifiedefficientfinetuning}.
For GRPO, we continue training from the multimodal SFT checkpoint using only the multimodal physics training set. We use \texttt{verl}~\citep{Sheng_2025} with LoRA rank 64, $\alpha=128$, and a learning rate of $5\times10^{-5}$ for 1 epoch. We evaluate open-ended physics reasoning using a hybrid evaluation framework, with details in Appendix~\ref{appendix:eval}.

\section{Evaluation Details}

\subsection{Evaluation Benchmarks}
\label{appendix:eval_benchmarks}
We evaluate our method on five multimodal physics reasoning benchmarks. PhysReason~\citep{zhang2025physreason}, PhyX~\citep{shen2025phyxdoesmodelwits}, and SeePhys~\citep{xiang2025seephys} are used for in-domain evaluation, while PhysUniBench~\citep{wang2026physunibenchmultimodalphysicsreasoning} and MMK-12$_{\mathrm{phys}}$~\citep{meng2025mmeurekaexploringfrontiersmultimodal} are used for out-of-domain evaluation. For the in-domain benchmarks, we hold out 1,000 problems for evaluation, including 467 from PhysReason, 412 from PhyX, and 121 from SeePhys. These problems are separated from the training pool before synthetic augmentation and physical modeling trace construction and are not used at any stage of training-data generation. We use the English multiple-choice subset of PhysUniBench and the physics subset of MMK-12, denoted as MMK-12$_{\mathrm{phys}}$.
\begin{table}[t]
\centering
\footnotesize
\setlength{\tabcolsep}{3pt}
\begin{tabular}{p{0.20\linewidth}p{0.74\linewidth}}
\toprule
\textbf{Model} & \textbf{Reasoning Excerpt} \\
\midrule
Baseline 
&
``By conservation of energy:
\[
(m_1 - m_2)gh + \frac{1}{2}(m_1+m_2)v^2 = 0
\]
Solve for $v^2$:
\[
v^2 = \frac{2(m_2-m_1)gh}{m_1+m_2}
\]
Wait --- this gives a negative value for $v^2$, which is impossible. That means we have the sign wrong. The net potential energy loss is $m_1gh-m_2gh$, which is positive, and this energy is converted to kinetic energy.''
\\
\midrule
SFT\\cold start
&
``The system consists of two buckets connected by a lightweight, inextensible rope passing over a frictionless, massless pulley. 
Since the rope is inextensible, both buckets move with the same magnitude of acceleration $a$ and reach the same final speed $v$.

Determine Acceleration: Apply Newton's Second Law to the system. The net force is the difference between the weights of the two buckets. Relate this net force to the total mass to solve for acceleration $a$.

\[
a=\frac{(m_1-m_2)g}{m_1+m_2}
\]''
\\
\bottomrule
\end{tabular}
\caption{Reasoning excerpts from an Atwood-machine example.}
\label{tab:reflective_case}
\end{table}

\subsection{Evaluation Protocol}
\label{appendix:eval}

Evaluating open-ended physics reasoning poses unique challenges due to the diversity of valid answer forms, including numerical values with units, symbolic expressions, and free-form textual responses. To ensure reliable and consistent evaluation, we adopt a hybrid evaluation framework that combines rule-based matching with LLM-assisted verification.

\paragraph{Answer Types.}
We categorize answers into four types: \emph{Numerical}, \emph{Symbolic}, \emph{Textual}, and \emph{Multiple-choice}. Each type is evaluated using a specialized strategy tailored to its structure.

\paragraph{Numerical Evaluation.}
For numerical answers, we first normalize both predicted and ground-truth responses into a canonical value–unit representation. This normalization handles diverse formats, including scientific notation, LaTeX expressions (e.g., fractions and square roots), and implicit unit expressions.
We then evaluate correctness based on (i) numerical equivalence under a relative tolerance, and (ii) unit consistency. When units differ but are dimensionally equivalent, we convert values to a shared canonical form before comparison (e.g., $\mathrm{eV} \leftrightarrow \mathrm{J}$, $\mathrm{deg} \leftrightarrow \mathrm{rad}$). The evaluation also supports multiple-value answers by matching predicted and ground-truth value–unit pairs.

\paragraph{Symbolic Evaluation.}
For symbolic answers, we normalize LaTeX expressions to remove formatting variations and parse them into symbolic representations. We then evaluate equivalence using the \texttt{math-verify} toolkit.\footnote{https://github.com/huggingface/Math-Verify}
To further improve robustness, we apply additional normalization steps such as removing presentation-level commands and, when applicable, comparing only the right-hand side of equations.

\paragraph{Multiple-Choice Evaluation.}
For multiple-choice questions, we extract the predicted option (e.g., A--E) using pattern matching and directly compare it with the ground-truth label.

\paragraph{Fallback LLM-as-a-Judge.}
When rule-based evaluation fails (e.g., due to parsing errors, unsupported formats, or ambiguous answers), we employ an LLM-as-a-judge as a fallback. The model is prompted to determine whether the predicted answer is equivalent to the ground truth and outputs a binary correctness decision. This mechanism ensures coverage for edge cases while preserving consistency with the primary evaluation criteria.

This hybrid design balances precision and robustness: rule-based evaluation provides strict and interpretable correctness signals, while the LLM-as-a-judge fallback handles the inherent variability of open-ended physics answers.

\section{Analysis of Reflective Reasoning Behavior}
\label{appendix:reflective_reasoning}
\paragraph{Reflection Keyword Detection.}
To quantify reflective reasoning behavior, we use a lightweight keyword-based diagnostic. 
A generated response is counted as containing reflection behavior if it includes at least one reflection-related expression, such as ``wait'', ``hold on'', ``let me check'', ``rethink'', or ``verify''. 
These keywords are intended to capture explicit self-checking or reconsideration patterns in the model's reasoning trace. 
We apply this diagnostic to 500 sampled Qwen3-VL-8B-Instruct generations and report the proportion of responses containing such reflection keywords.
As shown in Table~\ref{tab:reflection_behavior}, the baseline exhibits substantially more reflection-keyword patterns than Modeling-SFT, suggesting that the base model more frequently relies on explicit self-revision during reasoning.

\paragraph{Qualitative Analysis.}
Table~\ref{tab:reflective_case} shows a problem asks for the final speed of a two-bucket Atwood system released from rest, where the $12.0\,\mathrm{kg}$ bucket starts $2.00\,\mathrm{m}$ above the floor and is connected to a $4.0\,\mathrm{kg}$ bucket by a lightweight rope over a frictionless pulley. Both models solve the problem correctly and obtain $\boxed{4.43\,\mathrm{m/s}}$. However, the baseline first produces an incorrect intermediate sign convention and then recovers through an explicit reflective correction. In contrast, SFT cold start begins by constructing the physical system. This suggests that structured modeling can reduce the need for explicit self-revision by making the relevant physical constraints available before mathematical execution.

\clearpage
\onecolumn
\section{Qualitative Analysis}
\label{Appendix: Qualitative_Analysis}

\begin{questionbox}
\includegraphics[width=0.30\linewidth]{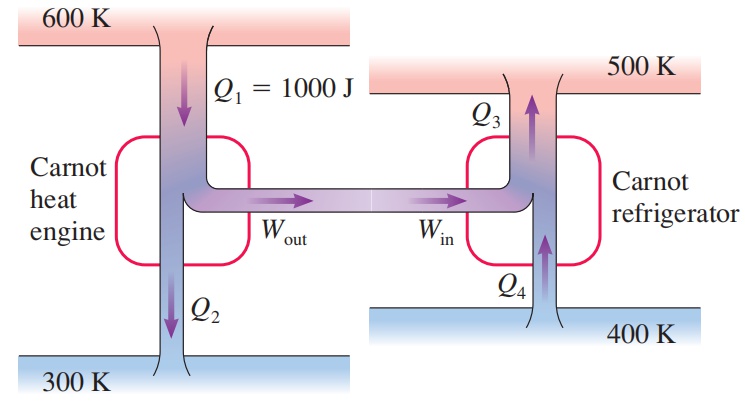}

\noindent
The figure shows a Carnot heat engine driving a Carnot refrigerator.
Determine $Q_3$.
\end{questionbox}

\begin{modelingbox}
\textbf{Objective:}

The problem involves a Carnot heat engine that supplies the work input to a Carnot refrigerator. The engine and refrigerator must therefore be treated as a coupled system. The first law of thermodynamics, Carnot efficiency, and the coefficient of performance of a Carnot refrigerator are used to determine the heat transfers.

\noindent
Goal: Determine the heat $Q_3$ rejected by the refrigerator to the $500\,\mathrm{K}$ reservoir.

\vspace{0.4em}
\textbf{Variables:}
\begin{itemize}
    \setlength{\itemsep}{2pt}
    \item $T_H=600\,\mathrm{K}$ \;|\; Hot-reservoir temperature of the heat engine
    \item $T_C=300\,\mathrm{K}$ \;|\; Cold-reservoir temperature of the heat engine
    \item $Q_1=1000\,\mathrm{J}$ \;|\; Heat absorbed by the heat engine
    \item $Q_2$ \;|\; Heat rejected by the heat engine
    \item $W_{\mathrm{out}}$ \;|\; Work output of the heat engine
    \item $T_{\mathrm{ref,hot}}=500\,\mathrm{K}$ \;|\; Hot-reservoir temperature of the refrigerator
    \item $T_{\mathrm{ref,cold}}=400\,\mathrm{K}$ \;|\; Cold-reservoir temperature of the refrigerator
    \item $W_{\mathrm{in}}$ \;|\; Work input to the refrigerator
    \item $Q_4$ \;|\; Heat absorbed by the refrigerator from the cold reservoir
    \item $Q_3$ \;|\; Heat rejected by the refrigerator to the hot reservoir
\end{itemize}

\vspace{0.4em}
\textbf{Constraints:}
\begin{enumerate}
    \setlength{\itemsep}{2pt}
    \item The heat engine and refrigerator operate as ideal, reversible Carnot machines.
    \item The engine work output is transferred to the refrigerator without loss, so $W_{\mathrm{in}}=W_{\mathrm{out}}$.
    \item Energy conservation for the refrigerator gives $Q_3=Q_4+W_{\mathrm{in}}$.
\end{enumerate}

\vspace{0.4em}
\textbf{Physical Modeling Process:}
\begin{enumerate}
    \setlength{\itemsep}{3pt}
    \item \textbf{Interpret the diagram and separate the two devices.}
    For the heat engine, $Q_1$ is absorbed from the $600\,\mathrm{K}$ reservoir and $Q_2$ is rejected to the $300\,\mathrm{K}$ reservoir. For the refrigerator, $Q_4$ is absorbed from the $400\,\mathrm{K}$ reservoir and $Q_3$ is rejected to the $500\,\mathrm{K}$ reservoir.
    \item \textbf{Model the heat engine.}
    Since it is a reversible Carnot engine, its efficiency and work output satisfy
    \[
    \eta_{\mathrm{eng}}=1-\frac{T_C}{T_H},
    \qquad
    W_{\mathrm{out}}=\eta_{\mathrm{eng}}Q_1.
    \]
    \item \textbf{Connect the two devices through work.}
    The engine drives the refrigerator directly, so the engine work output becomes the refrigerator work input:
    \[
    W_{\mathrm{in}}=W_{\mathrm{out}}.
    \]
    \item \textbf{Model the refrigerator.}
    Its cold-side heat is related to the work input through the Carnot coefficient of performance,
    \[
    \mathrm{COP}_{R}
    =\frac{Q_4}{W_{\mathrm{in}}}
    =\frac{T_{\mathrm{ref,cold}}}
    {T_{\mathrm{ref,hot}}-T_{\mathrm{ref,cold}}}.
    \]
    Energy conservation then gives $Q_3=Q_4+W_{\mathrm{in}}$.
    \item \textbf{Form the calculation sequence.}
    First obtain $W_{\mathrm{out}}$, pass it to the refrigerator as $W_{\mathrm{in}}$, determine $Q_4$, and finally calculate the requested $Q_3$.
\end{enumerate}

\vspace{0.6em}
\textbf{Execution:}

\textbf{Step 1: Calculate the engine work output.}
\[
\eta_{\mathrm{eng}}
=1-\frac{300}{600}
=0.5,
\qquad
W_{\mathrm{out}}
=\eta_{\mathrm{eng}}Q_1
=0.5(1000)
=500\,\mathrm{J}.
\]

\textbf{Step 2: Determine the refrigerator work input.}
\[
W_{\mathrm{in}}=W_{\mathrm{out}}=500\,\mathrm{J}.
\]

\textbf{Step 3: Calculate the refrigerator coefficient of performance.}
\[
\mathrm{COP}_{R}
=\frac{400}{500-400}
=4.
\]

\textbf{Step 4: Determine $Q_4$ and $Q_3$.}
\[
Q_4=\mathrm{COP}_{R}W_{\mathrm{in}}
=4(500)
=2000\,\mathrm{J},
\]
\[
Q_3=Q_4+W_{\mathrm{in}}
=2000+500
=2500\,\mathrm{J}.
\]

\vspace{0.4em}
\textbf{Answer:}
\[
\boxed{Q_3=2500\,\mathrm{J}}
\]
\end{modelingbox}

\begin{questionbox}
\includegraphics[width=0.24\linewidth]{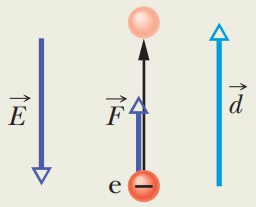}

\noindent
Near Earth's surface, a uniform electric field of magnitude
$E=150\,\mathrm{N/C}$ is directed downward. What is the change
$\Delta U$ in the electric potential energy of an electron when the
electric force moves it vertically upward through $d=520\,\mathrm{m}$?
\end{questionbox}

\begin{modelingbox}
\textbf{Objective:}

The problem asks for the change in electric potential energy of an electron moving upward in a uniform electric field. The force direction must first be determined from the sign of the electron charge. The work done by the electric field can then be related to the change in potential energy.

\noindent
Goal: Determine $\Delta U$ for the electron over the given displacement.

\vspace{0.4em}
\textbf{Variables:}
\begin{itemize}
    \setlength{\itemsep}{2pt}
    \item $E=150\,\mathrm{N/C}$ \;|\; Magnitude of the downward electric field
    \item $d=520\,\mathrm{m}$ \;|\; Upward displacement of the electron
    \item $q_e=-1.602\times10^{-19}\,\mathrm{C}$ \;|\; Electron charge
    \item $\Delta U$ \;|\; Change in electric potential energy
\end{itemize}

\vspace{0.4em}
\textbf{Constraints:}
\begin{enumerate}
    \setlength{\itemsep}{2pt}
    \item The electric field is uniform and directed downward.
    \item The electron moves vertically upward.
    \item Because the electron has negative charge, its electric force is opposite to the field and therefore points upward.
\end{enumerate}

\vspace{0.4em}
\textbf{Physical Modeling Process:}
\begin{enumerate}
    \setlength{\itemsep}{3pt}
    \item \textbf{Set the coordinate direction.}
    Choose upward as positive. Then
    $\vec{E}=-E\hat{\mathbf{y}}$ and
    $\vec{d}=d\hat{\mathbf{y}}$.
    \item \textbf{Infer the force from the charge sign.}
    The electron is negatively charged, so its force is opposite to the downward field:
    \[
    \vec{F}=q_e\vec{E}=eE\hat{\mathbf{y}}.
    \]
    The force therefore points upward, in the same direction as the displacement.
    \item \textbf{Determine the sign of the energy change.}
    Because the force and displacement are aligned, the electric field does positive work:
    \[
    W_{\mathrm{field}}=\vec{F}\cdot\vec{d}=eEd.
    \]
    The electron's electric potential energy must therefore decrease.
    \item \textbf{Form the executable relation.}
    Using $\Delta U=-W_{\mathrm{field}}$ gives
    \[
    \Delta U=-eEd.
    \]
\end{enumerate}

\vspace{0.6em}
\textbf{Execution:}

\textbf{Step 1: Determine the force direction and magnitude.}
\[
\vec{F}=q_e\vec{E}
=(-e)(-E\hat{\mathbf{y}})
=eE\hat{\mathbf{y}},
\]
\[
F=eE
=(1.602\times10^{-19})(150)
=2.403\times10^{-17}\,\mathrm{N}.
\]
The force and displacement are both upward, so the electric field does positive work.

\textbf{Step 2: Calculate the work done by the field.}
\[
W_{\mathrm{field}}
=eEd
=(1.602\times10^{-19})(150)(520)
=1.24956\times10^{-14}\,\mathrm{J}.
\]

\textbf{Step 3: Calculate the change in potential energy.}
\[
\Delta U
=-W_{\mathrm{field}}
=-1.24956\times10^{-14}\,\mathrm{J}
\approx-1.2\times10^{-14}\,\mathrm{J}.
\]

\vspace{0.4em}
\textbf{Answer:}
\[
\boxed{\Delta U\approx-1.2\times10^{-14}\,\mathrm{J}}
\]
\end{modelingbox}

\begin{questionbox}
\includegraphics[width=0.20\linewidth]{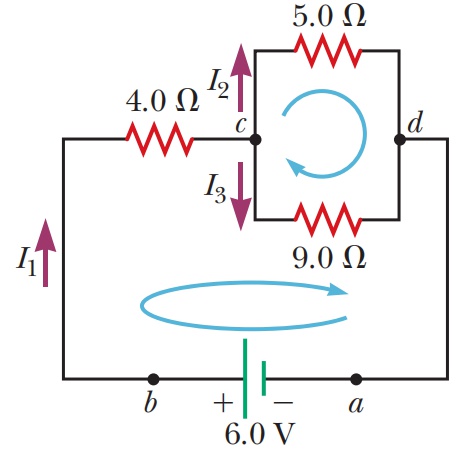}

\noindent
Find the current $I_1$ in the circuit shown in the figure using Kirchhoff's rules.
\end{questionbox}

\begin{modelingbox}
\textbf{Objective:}

The circuit contains a $4.0\,\Omega$ resistor in series with two parallel branches containing $5.0\,\Omega$ and $9.0\,\Omega$ resistors. Kirchhoff's current and voltage laws are used to relate the total current to the two branch currents.

\noindent
Goal: Determine $I_1$, the current supplied by the $6.0\,\mathrm{V}$ battery. The branch currents $I_2$ and $I_3$ are also calculated to verify the result.

\vspace{0.4em}
\textbf{Variables:}
\begin{itemize}
    \setlength{\itemsep}{2pt}
    \item $V=6.0\,\mathrm{V}$ \;|\; Battery voltage
    \item $R_1=4.0\,\Omega$ \;|\; Series resistance carrying $I_1$
    \item $R_2=5.0\,\Omega$ \;|\; Upper-branch resistance carrying $I_2$
    \item $R_3=9.0\,\Omega$ \;|\; Lower-branch resistance carrying $I_3$
    \item $I_1$ \;|\; Total current through the battery and $4.0\,\Omega$ resistor
    \item $I_2$ \;|\; Current through the $5.0\,\Omega$ branch
    \item $I_3$ \;|\; Current through the $9.0\,\Omega$ branch
\end{itemize}

\vspace{0.4em}
\textbf{Constraints:}
\begin{enumerate}
    \setlength{\itemsep}{2pt}
    \item The wires have negligible resistance and the battery maintains a constant voltage.
    \item At node $c$, the total current splits between the two branches, so $I_1=I_2+I_3$.
    \item The $5.0\,\Omega$ and $9.0\,\Omega$ resistors are connected across the same two nodes and have the same voltage drop.
\end{enumerate}

\vspace{0.4em}
\textbf{Physical Modeling Process:}
\begin{enumerate}
    \setlength{\itemsep}{3pt}
    \item \textbf{Identify the circuit topology.}
    The $4.0\,\Omega$ resistor carries the total current $I_1$ before the circuit splits into the $5.0\,\Omega$ and $9.0\,\Omega$ branches.
    \item \textbf{Apply current conservation at the junction.}
    At node $c$,
    \[
    I_1=I_2+I_3.
    \]
    \item \textbf{Construct the upper-loop equation.}
    Traversing the battery, the $4.0\,\Omega$ resistor, and the $5.0\,\Omega$ branch gives
    \[
    6.0-4.0I_1-5.0I_2=0.
    \]
    \item \textbf{Construct the lower-loop equation.}
    Traversing the battery, the $4.0\,\Omega$ resistor, and the $9.0\,\Omega$ branch gives
    \[
    6.0-4.0I_1-9.0I_3=0.
    \]
    \item \textbf{Form the solvable system.}
    The junction equation and the two loop equations provide three independent equations for $I_1$, $I_2$, and $I_3$.
\end{enumerate}

\vspace{0.6em}
\textbf{Execution:}

\textbf{Step 1: Express the branch currents in terms of $I_1$.}
\[
I_2=\frac{6.0-4.0I_1}{5.0},
\qquad
I_3=\frac{6.0-4.0I_1}{9.0}.
\]

\textbf{Step 2: Substitute into Kirchhoff's current law.}
\[
I_1
=\frac{6.0-4.0I_1}{5.0}
+\frac{6.0-4.0I_1}{9.0}.
\]
Multiplying by $45$ gives
\[
45I_1
=9(6.0-4.0I_1)+5(6.0-4.0I_1)
=84-56I_1.
\]
Therefore,
\[
101I_1=84,
\qquad
I_1=\frac{84}{101}\,\mathrm{A}
\approx0.832\,\mathrm{A}.
\]

\textbf{Step 3: Calculate the branch currents.}
\[
I_2
=\frac{6.0-4.0(84/101)}{5.0}
=\frac{54}{101}\,\mathrm{A}
\approx0.535\,\mathrm{A},
\]
\[
I_3
=\frac{6.0-4.0(84/101)}{9.0}
=\frac{30}{101}\,\mathrm{A}
\approx0.297\,\mathrm{A}.
\]
The values satisfy $I_1=I_2+I_3$.

\vspace{0.4em}
\textbf{Answer:}
\[
\boxed{I_1\approx0.83\,\mathrm{A}}
\]
\end{modelingbox}

\end{document}